\documentclass[sigconf]{acmart}

\usepackage{algorithm}
\usepackage{algorithmic}
\usepackage{amsmath}
\usepackage{multirow}
\usepackage{multicol}
\usepackage{balance}

\AtBeginDocument{%
  \providecommand\BibTeX{{%
    \normalfont B\kern-0.5em{\scshape i\kern-0.25em b}\kern-0.8em\TeX}}}

\setcopyright{acmcopyright}
\copyrightyear{2024}
\acmYear{2024}
\setcopyright{acmlicensed}\acmConference[WWW '24 Companion]{Companion Proceedings of the ACM Web Conference 2024}{May 13--17, 2024}{Singapore, Singapore}
\acmBooktitle{Companion Proceedings of the ACM Web Conference 2024 (WWW '24 Companion), May 13--17, 2024, Singapore, Singapore}
\acmDOI{10.1145/3589335.3648326}
\acmISBN{979-8-4007-0172-6/24/05}

\begin{document}

\title{Cache-Aware Reinforcement Learning in Large-Scale Recommender Systems}

\author{Xiaoshuang Chen}
\authornote{Both authors contributed equally to this research.}
\orcid{0000-0003-1267-1680}
\affiliation{%
  \institution{Kuaishou Technology}
  \city{Beijing}
  \country{China}
}
\email{chenxiaoshuang@kuaishou.com}

\author{Gengrui Zhang}
\authornotemark[1]
\affiliation{%
  \institution{Kuaishou Technology}
  \city{Beijing}
  \country{China}
}
\email{zhanggengrui@kuaishou.com}

\author{Yao Wang}
\affiliation{%
  \institution{Kuaishou Technology}
  \city{Beijing}
  \country{China}}
\email{wangyiyan@kuaishou.com}

\author{Yulin Wu}
\affiliation{%
  \institution{Kuaishou Technology}
  \city{Beijing}
  \country{China}
}
\email{wuyulin@kuaishou.com}

\author{Shuo Su}
\affiliation{%
  \institution{Kuaishou Technology}
  \city{Beijing}
  \country{China}
}
\email{sushuo@kuaishou.com}

\author{Kaiqiao Zhan}
\affiliation{%
  \institution{Kuaishou Technology}
  \city{Beijing}
  \country{China}
}
\email{zhankaiqiao@kuaishou.com}

\author{Ben Wang}\authornote{Corresponding author.}
\affiliation{%
  \institution{Kuaishou Technology}
  \city{Beijing}
  \country{China}
}
\email{wangben@kuaishou.com}
\renewcommand{\shortauthors}{Xiaoshuang Chen et al.}

\begin{abstract}
Modern large-scale recommender systems are built upon computation-intensive infrastructure and usually suffer from a huge difference in traffic between peak and off-peak periods. In peak periods, it is challenging to perform real-time computation for each request due to the limited budget of computational resources. The recommendation with a cache is a solution to this problem, where a user-wise result cache is used to provide recommendations when the recommender system cannot afford a real-time computation. However, the cached recommendations are usually suboptimal compared to real-time computation, and it is challenging to determine the items in the cache for each user. In this paper, we provide a cache-aware reinforcement learning (CARL) method to jointly optimize the recommendation by real-time computation and by the cache. We formulate the problem as a Markov decision process with user states and a cache state, where the cache state represents whether the recommender system performs recommendations by real-time computation or by the cache. The computational load of the recommender system determines the cache state. We perform reinforcement learning based on such a model to improve user engagement over multiple requests. Moreover, we show that the cache will introduce a challenge called critic dependency, which deteriorates the performance of reinforcement learning. To tackle this challenge, we propose an eigenfunction learning (EL) method to learn independent critics for CARL. Experiments show that CARL can significantly improve the users' engagement when considering the result cache. CARL has been fully launched in Kwai app, serving over 100 million users.
\end{abstract}




\maketitle
\section{Introduction}
Recently, reinforcement learning (RL) has drawn a growing attraction in recommender systems \cite{chen2019top,gao2022kuairec,zou2019reinforcement,cai2023reinforcing,xue2022resact, zhang2024unex}. RL-based recommendation methods treat users as the environment and the recommender system as the agent and then model the sequential interactions between users and the recommender system.  RL achieves success in improving users' long-term engagements, such as the total dwell time \cite{chen2019top, zhang2024unex} and the user retention \cite{cai2023reinforcing}.

\begin{figure}
    \centering
    \includegraphics[width=\columnwidth]{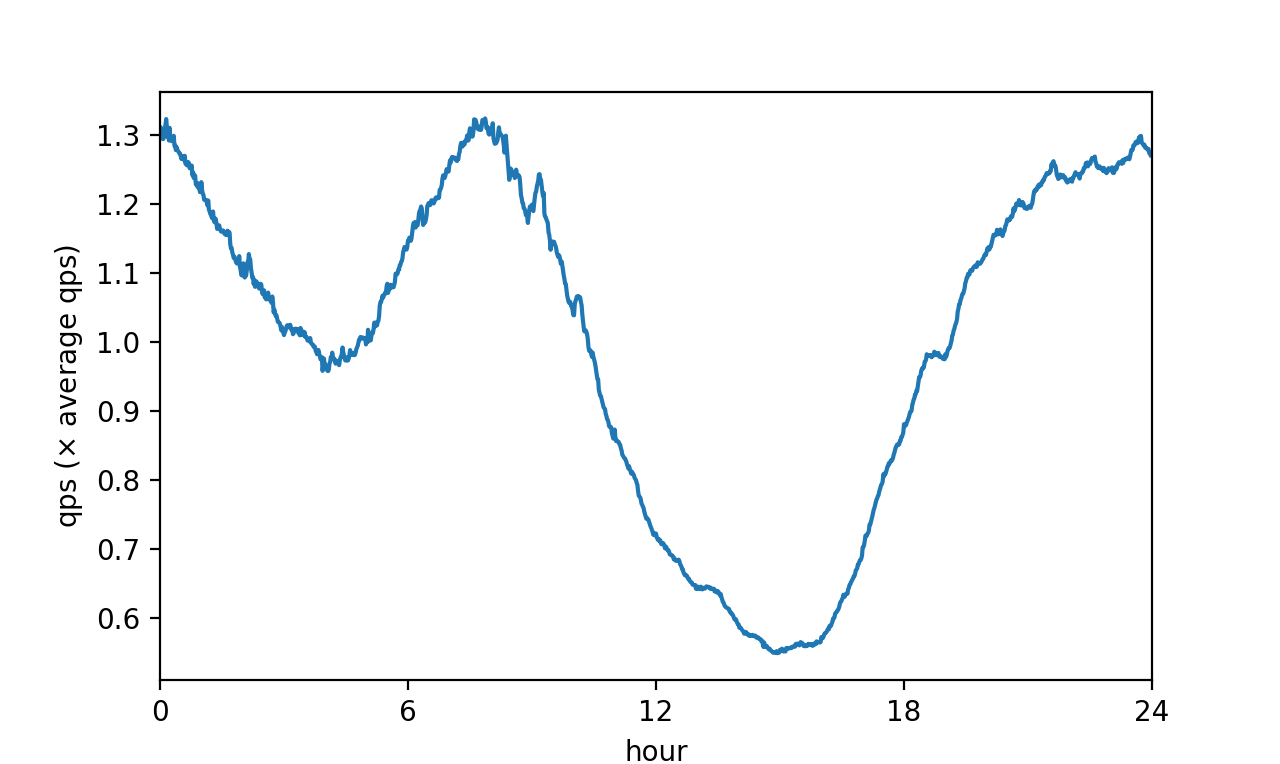}
    \caption{QPS in one day of Kwai app.}
    \label{fig:qps}
\end{figure}

\begin{figure}
    \centering
    \includegraphics[width=\columnwidth]{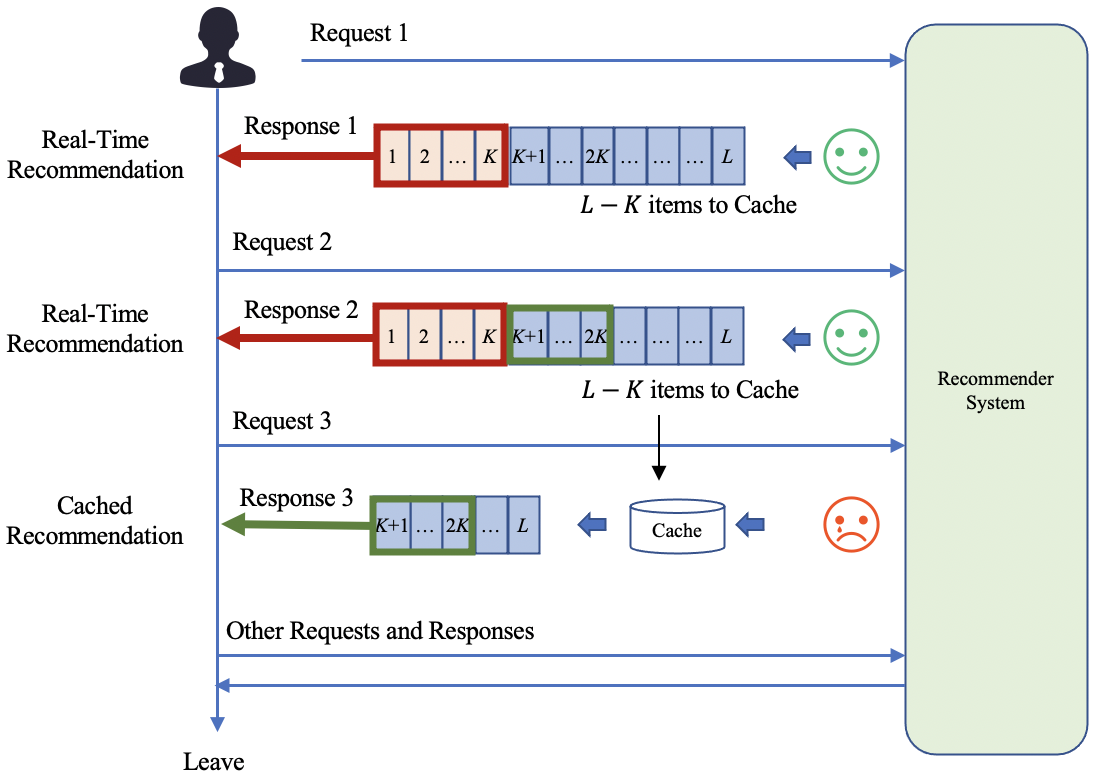}
    \caption{Recommendation with a result cache.}
    \label{fig:cache}
\end{figure}

Existing RL methods require the system to interact with the users and change the next recommendation according to the user feedback in real-time. However, modern large-scale recommender systems are built upon computation-intensive infrastructure \cite{wang2020cold}. Although there is an amount of research on reducing the computational burden of recommender systems \cite{jiang2020dcaf,johnson2019billion,liu2017cascade}, it is still challenging to perform real-time computation for each request during peak periods. Fig. \ref{fig:qps} shows the queries-per-second (QPS) of Kwai app based on the average QPS per day. It is shown that the computational burden during peak periods is several times that of off-peak periods. It requires a lot of computational resources to perform real-time computation in the peak period, while the same resources will be idle in the off-peak periods, which is not cost-effective. 
Therefore, a result cache \cite{wang2017caching,giannakas2018show} is used to balance the computational burden and the recommendation performance in large-scale recommender systems. Fig. \ref{fig:cache} briefly shows the recommender system with a result cache in Kwai. When a user sends a request, the system first provides a real-time recommendation, which returns a group of $L$ items, of which the top $K$ items are showed to the user while the other $L-K$ are put into a result cache. When the user's next request comes, the recommender system will perform a real-time computation if the traffic does not exceed the system's affordability, e.g. Response 2 in Fig. \ref{fig:cache}. Otherwise, it will recommend $K$ items directly from the result cache, e.g. Response 3 in Fig \ref{fig:cache}. The existence of the cache mitigates the computational burden of the recommender systems in peak periods, but it brings several challenges to traditional RL approaches:

\begin{itemize}
    \item \textbf{Lack of actions in the cache}. The RL module is usually a part of the real-time computation, meaning RL will not be performed when the request is processed by the cache. This contradicts the assumption of RL to interact with the users continuously. 
    \item \textbf{Unpredictability of the cache choice}. The choice of using real-time computation or the cache depends on the total computational burden of the system rather than the features of the specific request. Therefore, the cache choice is unpredictable by the RL algorithm, which increases the learning difficulty.
    \item \textbf{Heterogenuous rewards}. the performance of cached results is suboptimal compared to the results from the real-time computation due to the lack of instant behaviors of the user. Table \ref{table:cache-performance} provides an example in the Kwai app, where the average user engagement of cached recommendations is lower than that of real-time recommendations. Such characteristics need to be considered in the RL model.
\end{itemize}

\begin{table}[t]
\centering
\begin{tabular}{ccc}
    \toprule[1pt]
    \multirow{2}{*}{}
     & Real-Time & Cached  \\
     & Recommendations & Recommendations \\
    \hline
    Watch Time & $\times 1$ & $\times 0.85$ \\
    Like Rate & $\times 1$ & $\times 0.68$ \\
    Follow Rate  & $\times 1$ & $\times 0.54$ \\
    \bottomrule[1pt]
\end{tabular}
\caption{Comparisons between User engagement in real-time and cached recommendations in Kwai.}
\label{table:cache-performance}
\end{table}

To this end, we provide a novel cache-aware reinforcement learning (CARL) method. CARL introduces a cache state to model the choice of recommendations by real-time computation or by the cache. The cache state is determined by the computational burden of the recommender system and is independent of the user states and actions of the system. We also show that the existence of the cache introduces challenges to CARL training. Specifically, the uncontrollable cache state introduces an uncontrollable dependency between the critic functions of real-time cases and cached cases, which deteriorates the performance of the training algorithm. To tackle this challenge, we introduce a novel eigenfunction learning (EL) technique to train CARL. EL learns two independent critics and then combines them to obtain the critic functions of real-time and cached recommendations, improving CARL's performance.

In summary, our contributions are as follows: 
\begin{itemize}
    \item We introduce a novel CARL method to model the recommender systems with cache explicitly. 
    \item We show that the existence of the cache introduces critic dependency to the RL algorithm, which deteriorates the performance. Then, we provide an EL algorithm to train CARL effectively.
    \item Experiments show that CARL improves users' engagement in recommender systems with the cache. CARL has been launched in Kwai app, serving over 100 millions of users.
\end{itemize}

Following the introduction, Section \ref{sec:problem} provides the CARL model to describe the recommendation process with a cache. Section \ref{sec:learning} discusses the challenges of learning CARL, and provides an effective EL algorithm to train CARL. Section \ref{sec:experiment} provides the experimental results, and Section \ref{sec:conclusion} concludes the paper.

\section{Modeling of CARL} \label{sec:problem}
\begin{figure*}
    \centering
    \includegraphics[width=\textwidth]{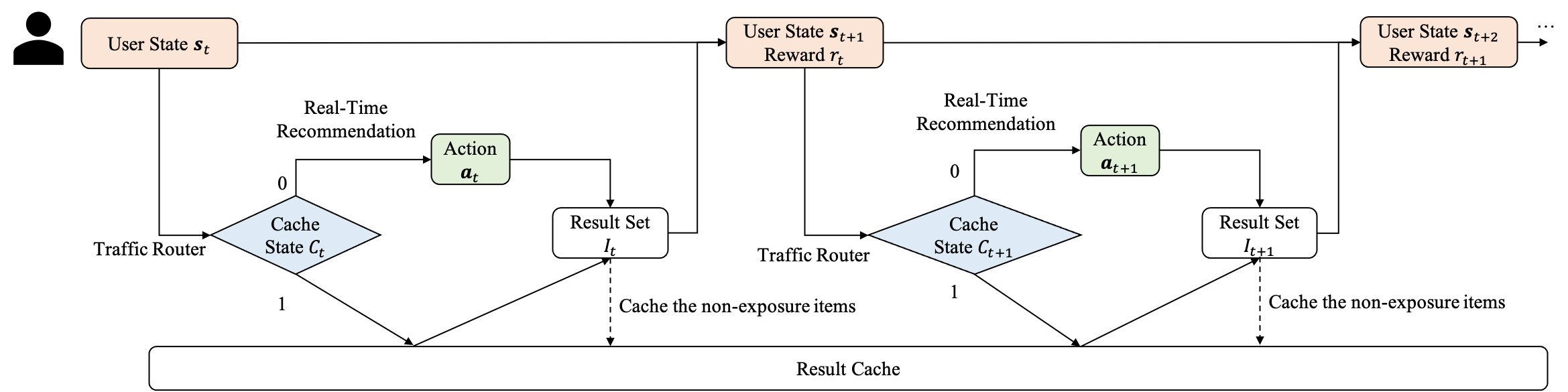}
    \caption{The CARL model.}
    \label{fig:carl}
\end{figure*}

This section provides the CARL model, which is an RL model considering the existence of the result cache, as shown in Figure \ref{fig:carl}. We model the interaction between users and the recommender systems as a Markov
Decision Process (MDP), where the recommender system is the agent, and
users are the environments.
When a user opens the app, a session begins, which consists of multiple requests until the user leaves the app. At Step $t$, the recommender system obtains a user state $\boldsymbol{s}_t$. The user state $\boldsymbol{s_t}$ consists of the user profile, behavior history, request context, and candidate
video features. According to the user request with the state $\boldsymbol{s}_t$, the recommender system generates an action $\boldsymbol{a}_t$, and outputs a set of items $I_t$ to the user according to the user state $\boldsymbol{s}_t$ and the action $\boldsymbol{a_t}$. After watching the provided items in $I_t$, the user provides feedback $r_t$, e.g. the watching time of the items and other interactions on the items. Then, the user transfers to the next state $\boldsymbol{s}_{t+1}$ and determines whether to send the next request or leave. The recommender system aims to maximize the long-term reward $R_t$ defined by
\begin{equation} \label{eq:r-t}
    \ R_t = \sum_{t'=t}^{T}\gamma^{t'-1}\mathbb{E}\left[r_{t'}\right],
\end{equation}
where $T$ is the last step and $\gamma$ is the discount factor.

A key component to be modeled in the recommender systems with cache is the choice of real-time computation or cached results. Specifically, the system maintains a result cache $\mathcal{K}_u$ for each user $u$. When a user request comes at Step $t$, a traffic router will provide a cache state $C_t\in\{0,1\}$, and the system will provide different services according to $C_t$:
\begin{itemize}
    \item \textbf{Recommendation by real-time computation}. When the recommender system has enough computational resources, the traffic router returns $C_t=0$, and the recommender system will use a real-time computation to provide recommended items. Specifically, $n$ deep models will be used to predict scores $\boldsymbol{x}_{j,t} \in \mathbb{R}^n$ for each candidate item $j$, in terms of various feedback (watch time, follow, like, etc.). Then, the action $\boldsymbol{a}_t$ is used to generate the final score of each item according to a parametrized ranking function $f$, i.e.
    \begin{equation}\label{eq:score_function}
        z_{j,t} = f(\boldsymbol{x}_{j,t};\boldsymbol{a}_t)
    \end{equation}
    Finally, the top $L$ items regarding the scores $z_{j,t}$ are used, of which the top $K$ items are recommended to the user, while the items with rank $K+1$ to $L$ are put into the result cache $\mathcal{K}_u$ for future use.
    \item \textbf{Recommendation by the cache}. When the traffic of the recommender system is too large to afford a real-time computation, the traffic router returns $C_t=1$, and the system will provide recommendations by the cache. In such cases, the recommender system obtains $K$ items directly from the user's result cache $\mathcal{K}_u$. Then, $\mathcal{K}_u$ is updated by removing the $K$ items recommended. There is no action $\boldsymbol{a}_t$ in recommendations by the cache.
\end{itemize}

In our scenario, the traffic router maintains a queue of ongoing real-time recommendation processes. When a new request comes, the traffic router checks the queue length. If the queue length does not exceed a given limit, it suggests a real-time computation; otherwise, it suggests a cached recommendation.

Given the abovementioned models, CARL aims to maximize the aggregate reward $R_t$ defined in Eq. \eqref{eq:r-t}.
The main challenge of training CARL arises from the stochastic characteristic of the traffic router. The cache state $C_t$ depends on the total traffic of the system rather than the user state $\boldsymbol{s}_t$. Therefore, $C_t$ is unpredictable according to $\boldsymbol{s}_t$. Such stochastic characteristics make it challenging to learn the critic. Section \ref{sec:learning} will discuss the challenge in detail and provide an effective training method.

\section{Learning of CARL} \label{sec:learning}
This section first shows the application of traditional RL algorithms to CARL, and discusses the critic dependency problem of the direct learning approach. Then, we provide the EL algorithm to learn CARL effectively.
\begin{table*}
\centering
\begin{tabular}{c|c|c}
    \toprule[1pt]
    &$C_t=0$&$C_t=1$ \\
    \hline
    $C_{t+1}=0$ &$\left[Q_0\left(\boldsymbol{s}_t,\boldsymbol{a}_t;\phi_0\right) - \left(r_t + \gamma Q_0\left(\boldsymbol{s}_{t+1},\mu\left(\boldsymbol{s}_{t+1};\theta^-\right);\phi_0^-\right)\right)\right]^2$ &$\left[Q_1\left(\boldsymbol{s}_t;\phi_1\right) - \left(r_t + \gamma Q_{0}\left(\boldsymbol{s}_{t+1},\mu\left(\boldsymbol{s}_{t+1};\theta^-\right);\phi_0^-\right)\right)\right]^2$ \\
    \hline
    $C_{t+1}=1$ & $\left[Q_0\left(\boldsymbol{s}_t,\boldsymbol{a}_t;\phi_0\right) - \left(r_t + \gamma Q_1\left(\boldsymbol{s}_{t+1},;\phi_1^-\right)\right)\right]^2$&$\left[Q_1\left(\boldsymbol{s}_t,\boldsymbol{a}_t;\phi_1\right) - \left(r_t + \gamma Q_1\left(\boldsymbol{s}_{t+1};\phi_1^-\right)\right)\right]^2$  \\
    \bottomrule[1pt]
\end{tabular}
\caption{Difference formulations of $l_{DL}(\phi)$ in Eq. \eqref{eq:dcl}}.
\label{table:dcl-cases}
\end{table*}
\subsection{Challenges of Direct Learning}
\subsubsection{The Direct Learning Algorithm}
We first show a direct application of traditional RL to learn the CARL model. We consider a typical actor-critic architecture. The action $\boldsymbol{a}_t$ is determined by a policy function $\mu$ with the parameter $\theta$:
\begin{equation} \label{eq:policy}
    \boldsymbol{a}_t = \mu\left(\boldsymbol{s}_t;\theta\right)
\end{equation}
We use a critic function $Q(\boldsymbol{s}_t, \boldsymbol{a}_t;\phi)$, parameterized by $\phi$ to estimate the long-term reward $R_t$ given the state $\boldsymbol{s}_t$ and the action $\boldsymbol{a}_t$. A typical critic learning algorithm \cite{lillicrap2015continuous} uses a temporal difference method to learn the parameter $\phi$ of the critic $Q$:
\begin{equation} \label{eq:ddpg-critic}
l\left(\phi\right) = \left[Q\left(\boldsymbol{s}_t,\boldsymbol{a}_t;\phi\right) - \left(r_t + \gamma Q\left(\boldsymbol{s}_{t+1},\mu\left(\boldsymbol{s}_{t+1};\theta^-\right);\phi^-\right)\right)\right]^2
\end{equation}
where $\theta^-$ is the parameter of the target actor, and $\phi^-$ is the parameter of the target critic. The critic function $Q$ is used as the reward function of the policy function $\mu$, and the parameter $\theta$ of the actor updates according to the following gradient ascent:
\begin{equation} \label{eq:ddpg-actor}
\nabla_\theta J = \nabla_\theta\mu(\boldsymbol{s}_t;\theta)\nabla_\mu Q\left(\boldsymbol{s}_t, \mu(\boldsymbol{s}_t;\theta)\right)
\end{equation}

To apply Eq. \eqref{eq:ddpg-critic}\eqref{eq:ddpg-actor} to CARL, we define the conditional critic functions: 
\begin{equation}
Q_0\left(\boldsymbol{s}_t,\boldsymbol{a}_t;\phi_0\right) \triangleq \mathbb{E}\left[R_t|C_t = 0\right],  Q_1\left(\boldsymbol{s}_t;\phi_1\right) \triangleq \mathbb{E}\left[R_t|C_t = 1\right]
\end{equation}
$Q_0$ and $Q_1$ are the expected long-term rewards under the condition that the Request $t$ is processed by real-time computation and by the cache, respectively. The input of $Q_1$ does not contain the action $\boldsymbol{a}_t$ because there is no action in recommendations by the cache.

Given $Q_0$ and $Q_1$, the total critic function $Q$ can be written as
\begin{equation} \label{eq:dl-critic}
\begin{aligned}
    Q(\boldsymbol{s}_t,\boldsymbol{a}_t;\phi) =Q_{C_t}(\boldsymbol{s}_t,\boldsymbol{a}_t;\phi_{C_t}), C_t \in \{0,1\}
\end{aligned}
\end{equation}
where $\phi = \{\phi_0,\phi_1\}$.
With the critic defined in Eq. \eqref{eq:dl-critic}, we can directly train the CARL by the following critic loss:

\noindent \textbf{CARL-DL (Direct Learning of CARL)}:
\begin{equation} \label{eq:dcl}
l_{DL}\left(\phi\right) = \left[Q_{C_t}\left(\boldsymbol{s}_t,\boldsymbol{a}_t;\phi\right) - \left(r_t + \gamma Q_{C_{t+1}}\left(\boldsymbol{s}_{t+1},\mu\left(\boldsymbol{s}_{t+1};\theta^-\right);\phi^-\right)\right)\right]^2,
\end{equation}
with the policy loss remains the same as Eq. \eqref{eq:ddpg-actor}.

\subsubsection{Critic Dependency}
\begin{figure}
    \centering
    \includegraphics[width=\columnwidth]{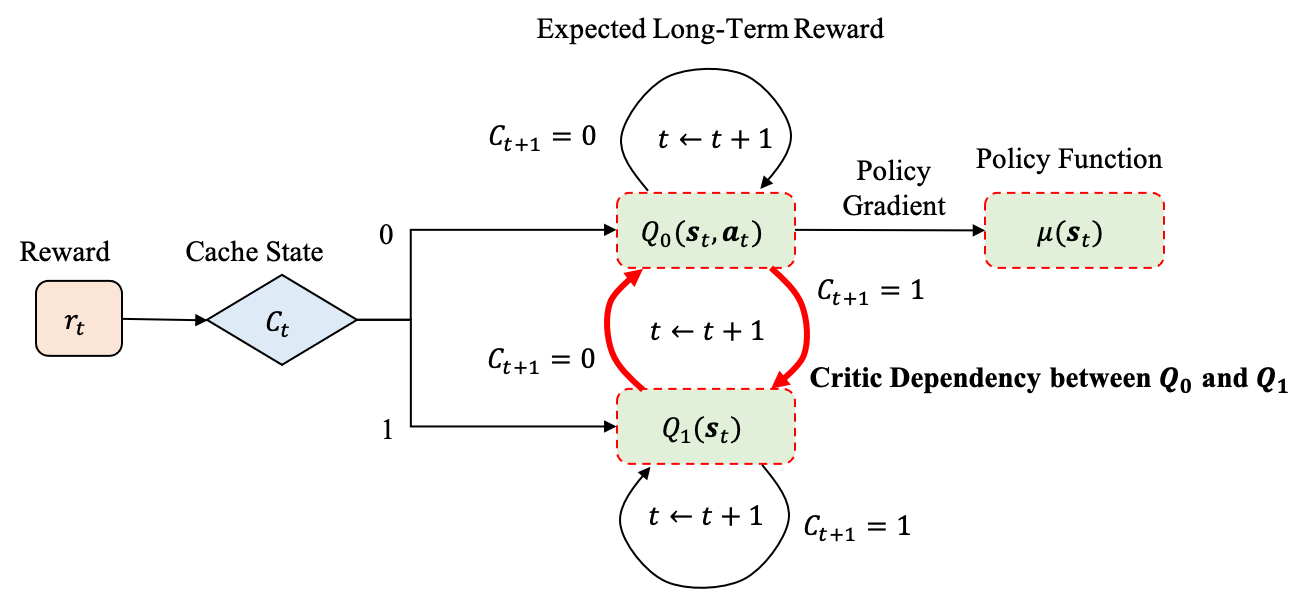}
    \caption{Direct Learning of CARL}
    \label{fig:dcl}
\end{figure}

The CARL-DL approach is straightforward, but it faces a considerable challenge, i.e. the \textbf{Critic Dependency}. Specifically, Eq. \eqref{eq:dcl} shows the dependency of $Q(\boldsymbol{s}_t,\boldsymbol{a}_t;\phi)$ and $Q(\boldsymbol{s}_{t+1},\boldsymbol{a}_{t+1};\phi)$. According to different cache states $C_t$ and $C_{t+1}$, there are four possible formulations of Eq. \eqref{eq:dcl}, as shown in Table \ref{table:dcl-cases}. It is clear that the $Q_0$ and $Q_1$ depend on each other. Figure \ref{fig:dcl} also shows the dependency among the two critics $Q_0$ and $Q_1$ in the DL of CARL. The critic dependency makes the learning process depending on the values of the cache states $C_t$ and $C_{t+1}$, but the cache states are highly stochastic, and are unpredictable by the states $\boldsymbol{s}_t$ and the actions $\boldsymbol{a}_t$. Such problem deteriorates the convergence of the critic learning.

To discuss this problem more precisely, we consider the transit function of the critics. Since the cache state $C_t$ is a stochastic variable, we use $D_c(t)$ to denote the probability that $C_t = c$:
\begin{equation}
    D_c(t) \triangleq P(C_t = c), c\in\{0,1\}
\end{equation}
$D_c(t)$ does not depend on the user's state $\boldsymbol{s}_t$ because it simply relies on the total requests per second and the maximal real-time recommendations per second.

In addition, we define the expectation of the immediate reward $r_t$ of different cache states $C_t$:
\begin{equation}
V_0\left(\boldsymbol{s}_t,\boldsymbol{a}_t\right) \triangleq \mathbb{E}\left[r_t|C_t = 0\right], V_1\left(\boldsymbol{s}_t\right) \triangleq \mathbb{E}\left[r_t|C_t = 1\right]
\end{equation}
Then, according to the backward induction, the transit functions of the expected immediate reward and the expected long-term reward can be written as:
\begin{equation} \label{eq:iteration}
\begin{aligned}
Q_0\left(\boldsymbol{s}_t,\boldsymbol{a}_t\right) = &V_0\left(\boldsymbol{s}_t,\boldsymbol{a}_t\right) + \gamma D_0\left(t+1\right)\mathbb{E}_{\boldsymbol{a}_{t+1}}Q_0\left(\boldsymbol{s}_{t+1},\boldsymbol{a}_{t+1}\right) \\
&+ \gamma D_1\left(t+1\right)\mathbb{E}_{\boldsymbol{a}_{t+1}}Q_1\left(\boldsymbol{s}_{t+1}\right) \\
Q_1\left(\boldsymbol{s}_t\right) =& V_1\left(\boldsymbol{s}_t\right) + \gamma D_0\left(t+1\right)\mathbb{E}_{\boldsymbol{a}_{t+1}}Q_0\left(\boldsymbol{s}_{t+1},\boldsymbol{a}_{t+1}\right) \\
&+ \gamma D_1\left(t+1\right)\mathbb{E}_{\boldsymbol{a}_{t+1}}Q_1\left(\boldsymbol{s}_{t+1}\right) \\
\end{aligned}
\end{equation}
Now, we provide more explanations on Eq. \eqref{eq:iteration}, as shown in Figure \ref{fig:dcl}. On the one hand, the critic functions $Q_0$ and $Q_1$, i.e. the expectation of the long-term reward $R_t$, depends on the immediate reward $r_t$, estimated by $V_0$ and $V_1$. On the other hand,  $Q_0$ and $Q_1$ also depend on the future feedback of the user. However, due to the stochastic characteristic of the cache state $C_t$, the future feedback of the user obeys different distributions given different $C_t$, and the expectation of the future feedback is described by the second and third term of the right part of Eq. \eqref{eq:iteration}.

Eq. \eqref{eq:iteration} also shows the critic dependency problem, since the $Q_0$ and $Q_1$ clearly depend on each other. Such dependency, together with the stochasticity of the cache states $C_t$, will increase the learning difficulty of CARL.
We will discuss the solution to the critic dependency problem in the following subsections.

\subsection{Eigenfunctions} \label{sec:eigen}
\begin{figure*}
    \centering
    \includegraphics[width=\textwidth]{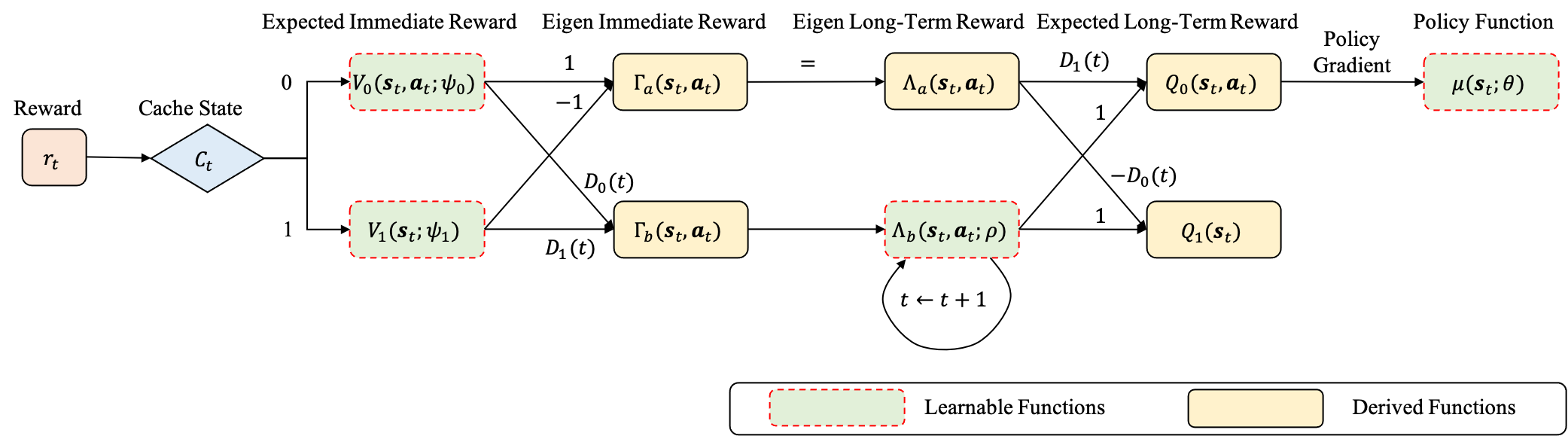}
    \caption{Eigenfunction Learning of CARL}
    \label{fig:el}
\end{figure*}
This paper uses the eigenfunction technique to solve the critic dependency problem. The motivation is to find a group of independent variables derived from the critic functions $Q_0$ and $Q_1$ so that the learning process of different critic functions can be decoupled.

We first define the eigen immediate reward and the eigen long-term reward.

\begin{definition}[Eigen Immediate Reward]
The eigen immediate rewards, denoted by $\Gamma_a$ and $\Gamma_b$, are given by
\begin{equation} \label{eq:gamma}
\begin{aligned}
\Gamma_a\left(\boldsymbol{s}_t,\boldsymbol{a}_t\right) &\triangleq V_0\left(\boldsymbol{s}_t,\boldsymbol{a}_t\right) - V_1\left(\boldsymbol{s}_t\right) \\
\Gamma_b\left(\boldsymbol{s}_t,\boldsymbol{a}_t\right) &\triangleq D_0(t)V_0\left(\boldsymbol{s}_t,\boldsymbol{a}_t\right) +D_1(t) V_1\left(\boldsymbol{s}_t\right)
\end{aligned}
\end{equation}
\end{definition}
The eigen immediate rewards $\Gamma_a$ and $\Gamma_b$ are the linear combinations of the expected immediate rewards $V_0$ and $V_1$. Similarly, we have
\begin{definition}[Eigen Long-Term Reward]
The eigen long-term rewards, denoted by $\Lambda_a$ and $\Lambda_b$, are given by
\begin{equation} \label{eq:lambda}
\begin{aligned}
\Lambda_a\left(\boldsymbol{s}_t,\boldsymbol{a}_t\right) &\triangleq Q_0\left(\boldsymbol{s}_t,\boldsymbol{a}_t\right) - Q_1\left(\boldsymbol{s}_t\right) \\
\Lambda_b\left(\boldsymbol{s}_t,\boldsymbol{a}_t\right) &\triangleq D_0(t)Q_0\left(\boldsymbol{s}_t,\boldsymbol{a}_t\right) +D_1(t) Q_1\left(\boldsymbol{s}_t\right)
\end{aligned}
\end{equation}
\end{definition}
Clearly, given the eigen long-term rewards $\Lambda_a$ and $\Lambda_b$, we can recover the expected long-term rewards $Q_0$ and $Q_1$ simply by solve the equations in Eq. \eqref{eq:lambda}. Specifically, we have
\begin{equation}\label{eq:recover-l}\\
\begin{aligned}
    Q_0(\boldsymbol{s}_t, \boldsymbol{a}_t) &= D_1(t)\Lambda_a(\boldsymbol{s}_t, \boldsymbol{a}_t) + \Lambda_b(\boldsymbol{s}_t, \boldsymbol{a}_t) \\
    Q_1(\boldsymbol{s}_t) &= -D_0(t)\Lambda_a(\boldsymbol{s}_t, \boldsymbol{a}_t) + \Lambda_b(\boldsymbol{s}_t, \boldsymbol{a}_t) 
\end{aligned}
\end{equation}
Therefore, once we can estimate the eigen long-term rewards $\Lambda_a$ and $\Lambda_b$, we can also estimate the expected long-term rewards $Q_0$ and $Q_1$.

Although the eigen immediate/long-term rewards are just equivalent transformations of the expected immediate/long-term rewards, their iterative functions are much simplified. Actually, we have
\begin{proposition} \label{prop:eigen}[Iterative Function of Eigen Long-Term Rewards]
The iterative function of the eigen long-term rewards writes
\begin{equation} \label{eq:eigen-iterate}
\begin{aligned}
\Lambda_a\left(\boldsymbol{s}_t,\boldsymbol{a}_t\right) &= \Gamma_a\left(\boldsymbol{s}_t,\boldsymbol{a}_t\right) \\
\Lambda_b\left(\boldsymbol{s}_t,\boldsymbol{a}_t\right) &= \Gamma_b\left(\boldsymbol{s}_t,\boldsymbol{a}_t\right) + \gamma \Lambda_b\left(\boldsymbol{s}_{t+1},\boldsymbol{a}_{t+1}\right)
\end{aligned}
\end{equation}
\end{proposition}
The proof can be referred to in Appendix. \ref{appendix:proof-eigen}. We explain more about Proposition \ref{prop:eigen}. Firstly, $\Lambda_a$ is the difference between the long-term rewards under recommendations by real-time computation and by the cache, i.e., $Q_0$ and $Q_1$. The first equation of Eq. \eqref{eq:eigen-iterate} shows that such a difference equals the difference between the immediate rewards under real-time and cached recommendations. It is because the future cache states are stochastic and independent of the current cache state. Secondly, $\Lambda_b$, which is the weighted sum of real-time and cached recommendations, follows an iterative equation independent of $\Lambda_a$.

Compared to the iterative functions of the expected long-term rewards in Eq. \eqref{eq:iteration} where the two critics depend on each other, the eigen long-term rewards $\Lambda_a$ and $\Lambda_b$ are independent. Therefore, if we regard the eigen long-term rewards as the function to learn, we will not suffer from the critic dependency problem.

Given the abovementioned discussions, we are ready to provide the final algorithm.

\subsection{The Eigenfunction Learning Algorithm}
Now, we provide the CARL-EL algorithm according to the eigenfunctions defined in Section \ref{sec:eigen}. The framework of CARL-EL is shown in Figure \ref{fig:el}. Specifically, we regard the following functions as learnable functions:
\begin{itemize}
    \item The immediate reward functions $V_0(\boldsymbol{s}_t,\boldsymbol{a}_t;\psi_0)$ and $V_1(\boldsymbol{s}_t;\psi_1)$, parameterized by $\psi_0$ and $\psi_1$.
    \item The eigen long-term reward function $\Lambda_b(\boldsymbol{s}_t,\boldsymbol{a}_t;\rho)$, parameterized by $\rho$.
    \item The policy function $\mu(\boldsymbol{s}_t;\theta)$, parameterized by $\theta$.
\end{itemize}
In contrast, we regard the following functions as derived functions from the abovementioned learnable functions:
\begin{itemize}
    \item The eigen immediate reward functions $\Gamma_a$ and $\Gamma_b$, which can be obtained from the expected immediate reward $V_0$ and $V_1$ according to Eq. \eqref{eq:gamma}.
    \item The eigen long-term reward function $\Lambda_a$, which is equal to $\Gamma_a$ according to Proposition \ref{prop:eigen}.
    \item The expected long-term rewards $Q_0$ and $Q_1$, which can be derived from the eigen long-term rewards $\Lambda_a$ and $\Lambda_b$ according to Eq. \eqref{eq:recover-l}.
\end{itemize}

The EL algorithm, as shown in Algorithm \ref{alg:training}, consists of five steps, i.e. learning the immediate rewards, calculating the eigen immediate rewards, learning the eigen long-term reward, recovering the long-term rewards, and learning the policy function.

\begin{algorithm*}
\caption{CARL-EL: Eigenfunction Learning of Cache-Aware Reinforcement Learning}
\label{alg:training}
\begin{algorithmic}[1]
\STATE Input: $\left\{\boldsymbol{s}_{1:T},\boldsymbol{a}_{1:T}, r_{1:T}\right\}$ for each user.
\STATE Output: A policy function $\mu\left(\boldsymbol{s}_t;\theta\right)$ parameterized by $\theta$.
\FOR{each user session with $T$ requests from the replay buffer}
    \FOR{$t = 1,\cdots, T$}
        \STATE Collect the user state $\boldsymbol{s}_t$, the reward $r_t$, and the action $\boldsymbol{a}_t$ from the replay buffer.
        \STATE \textbf{Step 1}: Learn the immediate rewards $V_0(\boldsymbol{s}_t, \boldsymbol{a}_t; \psi_0)$ and $V_1(\boldsymbol{s}_t;\psi_1)$ according to Eq. \eqref{eq:immediate-reward-learning}.
        \STATE \textbf{Step 2}: Calculate the eigen immediate rewards $\Gamma_a(\boldsymbol{s}_t, \boldsymbol{a}_t)$ and $\Gamma_b(\boldsymbol{s}_t, \boldsymbol{a}_t)$ according to Eq. \eqref{eq:gamma}. \\
        \STATE \textbf{Step 3}: Calculate the eigen long-term rewards $\Lambda_a(\boldsymbol{s}_t, \boldsymbol{a}_t)$, and learn $\Lambda_b(\boldsymbol{s}_t, \boldsymbol{a}_t;\rho)$ according to Eq. \eqref{eq:eigen-td}.
        \STATE \textbf{Step 4}: Calculate the critic function $Q_0(\boldsymbol{s}_t, \boldsymbol{a}_t)$ according to Eq. \eqref{eq:recover-l}.
        \STATE \textbf{Step 5}: Take policy gradient with the policy loss $Q_0(\boldsymbol{s}_t,\mu(\boldsymbol{s}_t;\theta))$ to update $\theta$ according to Eq. \eqref{eq:ddpg-actor}.
    \ENDFOR
\ENDFOR
\end{algorithmic}
\end{algorithm*}

\textbf{Step 1}: we learn the immediate reward functions $V_0(\boldsymbol{s}_t;\boldsymbol{a}_t; \psi_0)$ and $V_1(\boldsymbol{s}_t;\psi_1)$. The loss function is
\begin{equation} \label{eq:immediate-reward-learning}
    l({\psi}) = \left\{
    \begin{matrix}
    \left[r_t - V_0(\boldsymbol{s}_t, \boldsymbol{a}_t; \psi_0)\right]^2&, C_t = 0 \\
    \left[r_t - V_1(\boldsymbol{s}_t; \psi_1)\right]^2&, C_t = 1
    \end{matrix}
    \right.
\end{equation}
We learn different immediate rewards for requests processed by real-time and cached recommendations, where $C_t = 0$ or $1$, respectively.

\textbf{Step 2}: after obtaining the estimation of immediate value functions, we calculate the eigen immediate rewards $\Gamma_a(\boldsymbol{s}_t,\boldsymbol{a}_t)$ and $\Gamma_b(\boldsymbol{s_t},\boldsymbol{a}_t)$ according to Eq. \eqref{eq:gamma}.

\textbf{Step 3}: the eigen immediate rewards $\Gamma_a$ and $\Gamma_b$ are used to learn the eigen long-term rewards $\Lambda_a$ and $\Lambda_b$ via the temporal difference function. According to Proposition \ref{prop:eigen}, we have $\Lambda_a = \Gamma_a$, therefore, there is no need to learn $\Lambda_a$. In contrast, $\Lambda_b$ needs to be learned by the following temporal difference loss:
\begin{align} \label{eq:eigen-td}
l(\rho)&= \left[\Lambda_b\left(\boldsymbol{s}_t, \boldsymbol{a}_t;\rho\right) - \left(\Gamma_b(\boldsymbol{s}_t,\boldsymbol{a}_t) + \gamma \Lambda_b\left(\boldsymbol{s}_{t+1},\mu(\boldsymbol{s}_{t+1};\theta^-); \rho^-\right)\right)\right]^2
\end{align}
where $\theta^-$ and $\phi^-$ are the parameters of the target actor and target critic, respectively.

\textbf{Step 4}: we recover the expected long-term reward $Q_0$ and $Q_1$ from the eigen long-term reward $\Lambda_a$ and $\Lambda_b$ according to Eq. \eqref{eq:recover-l}.

\textbf{Step 5}: we regard the expected long-term reward $Q_0(\boldsymbol{s}_t,\boldsymbol{a}_t)$ as the policy loss, and take the policy gradient algorithm according to Eq. \eqref{eq:ddpg-actor}.

Figure \ref{fig:el} shows the information flow of Algorithm \ref{alg:training}. Compared with the direct learning method shown in Figure \ref{fig:dcl}, the main difference between Algorithm \ref{alg:training} and CARL-DL is the temporal difference equation, i.e. Eq. \eqref{eq:eigen-td} in CARL-EL and Eq. \eqref{eq:dcl} in CARL-DL. In CARL-DL, the learning of the critic functions $Q_0$ and $Q_1$ depend on each other according to a stochastic cache state $C_t$; while in CARL-EL, the learning of the eigen long-term rewards $\Lambda_a$ and $\Lambda_b$ does not depend on each other. Therefore, the temporal difference of CARL-EL does not rely on the stochastic cache state $C_t$, and hence, the variance of learning can be effectively reduced, and the performance can be improved. 

\section{Experimental Results} \label{sec:experiment}

We deploy our proposed CARL model in Kwai, a short video platform serving over 100 million users. The QPS is shown in Figure \ref{fig:qps}, and the traffic router executes as described in Section \ref{sec:problem}. The ratio of cached recommendations during peak periods is about 40\%. We did not conduct offline experiments because there is, to the best of the authors' knowledge, no offline experimental environment considering the impact of the cache for the time being.

\subsubsection{Implementation}
\begin{figure}
    \centering
    \includegraphics[width=\columnwidth]{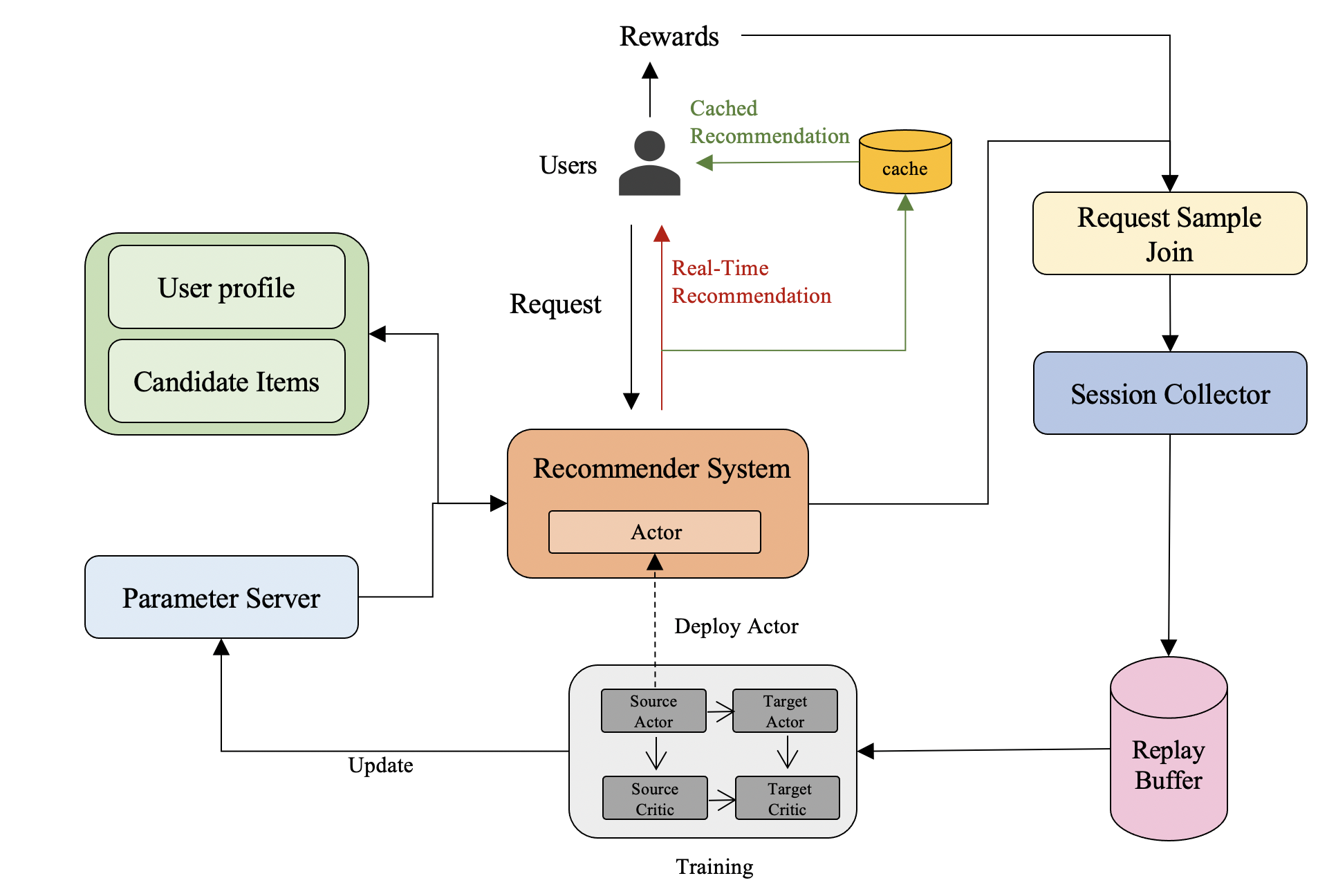}
    \caption{System Implementation}
    \label{fig:implementation}
\end{figure}

The structure of the online system is shown in Figure \ref{fig:implementation}, and the details are as follows:
\begin{itemize}
    \item \textbf{Sample Generation}. We first collect the states, actions, and rewards of each request to generate request-wise samples. Next, a session collector groups the requests of the same user and orders them by the timestamp of the requests. We split the request sample list into multiple sessions based on the 15-minute inactivity rule. The session collector constructs two consecutive requests from the same session into a single RL sample and feeds them into the replay buffer. 
    \item \textbf{MDP}. We deploy the actor in the ranking part of the recommender system. The recommender serves as the agent, and the users serve as the environment. At each user request, the actor gets the state $\boldsymbol{s}_t$ containing a vector of the user profile, the behavior history, the request context, and the statistics of candidate video features. The user profile covers the information collected in the registration, including the gender, the age, and the interests of the user. The behavior history includes the items that the user interacted with in the past. The request context includes the timestamp and the location, and the video statistics include the average score and the 10-th/30-th/50-th/70-th/90-th percentile of the prediction scores. Then, it returns an action vector $\boldsymbol{a}_t$, which is a 5-dimensional continuous vector ranging in $[0, 3]$, acting as the fusion parameters of five scoring models predicting the main feedback (watchtime, shortview, longview, finish, and forward). A final ranking score is generated according to Eq. \eqref{eq:score_function}, of which the top $L$ items are selected, and the first $K$ items are shown to the users (the red arrow in Figure \ref{fig:implementation}). The rest $L-K$ items are cached in case of a cached recommendation when the traffic exceeds the system’s affordability(the green arrow in Figure \ref{fig:implementation}). In our scenario, $L=40$ and $K=8$.
\end{itemize}

\label{sec:live}
\subsection{Settings}

\subsubsection{Baselines}
We deploy different methods in the online system:
\begin{itemize}
    \item \textbf{Cross Entropy Method (CEM)} \cite{rubinstein2004cross}: A black-box optimization method commonly used for hyper-parameter optimization. We use CEM to search the best parameters $\boldsymbol{a}_t$.
    \item \textbf{TD3} \cite{fujimoto2018addressing}: A famous RL method which uses twin critics to reduce the bias. TD3 does not explicitly distinguish the real-time and cached recommendations.
    \item \textbf{RLUR}\cite{cai2023reinforcing}: A state-of-the-art reinforcement-learning-based method for the fusion of multiple predictions. RLUR is the last online version before CARL is deployed in Kwai. Similar to TD3, RLUR does not explicitly distinguish the real-time and cached recommendations, either.
    \item \textbf{CARL-DL}: The CARL with the direct learning method.
    \item \textbf{CARL-EL}: The CARL with the eigenfunction learning method.
\end{itemize}

\subsubsection{Metrics}
We evaluate the abovementioned algorithms by the session-wise watch time and the daily watch time, an important metric in short video platforms \cite{zhan2022deconfounding}. We show the experiment results in improvement compared to the CEM method.

\subsection{Results}
Table \ref{table:results} shows the comparison results of different methods in online A/B experiments. In our scenario, even a 0.1\% improvement in the watch time is significant. It is shown that RLUR and TD3 outperform CEM due to the effectiveness of RL approaches. Both the CARL-DL and the CARL-EL outperform RLUR and TD3, showing the effectiveness of the cache-aware modeling. Moreover, the CARL-EL achieves the best performance, showing the effectiveness of the proposed EL method.

\begin{table}
\centering
\begin{tabular}{ccc}
    \toprule[1pt]
     & Session&  Daily \\
     & Watch Time & Watch Time \\
    \hline
    CEM & +0.000\% & +0.000\% \\
    \hline
    TD3 & +0.176\%  & +0.184\%  \\
    RLUR & +0.206\% &  +0.192\%  \\
    CARL-DL & +0.390\%   & +0.342\%  \\
    CARL-EL  &\textbf{+0.545\%}  & \textbf{+0.586\%} \\
    \bottomrule[1pt]
\end{tabular}
\caption{Experimental Results in Kwai in terms of the improvement over CEM.}
\label{table:results}
\end{table}

\subsection{Discussions}
\subsubsection{Improvement on Critic Loss}
Table \ref{table:critic-loss} shows the average critic loss of different RL-based methods over one day. We evaluate these methods' critic loss using the formula shown in Eq. \eqref{eq:ddpg-critic}. It is shown that CARL outperforms the other algorithms, showing the effectiveness of the explicit modeling of cache. Moreover, CARL-EL achieves a better critic loss than CARL-DL, showing that the EL approach effectively improves the performance of CARL.
\begin{table}
\centering
\begin{tabular}{ccccc}
    \toprule[1pt]
    Methods & TD3 & RLUR & CARL-DL & CARL-EL \\
    \hline
    Loss & 0.33  & 0.30  & 0.28  & \textbf{0.25} \\
    \bottomrule[1pt]
\end{tabular}
\caption{Critic Loss of Different Methods.}
\label{table:critic-loss}
\end{table}

\subsubsection{Gaps between Recommendations by Real-Time Computation and by the Cache}
\begin{figure}
    \centering
    \includegraphics[width=\columnwidth]{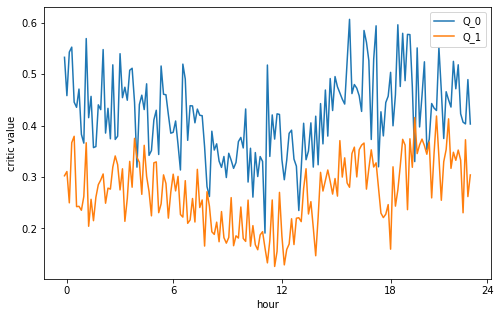}
    \caption{Comparisons of critic values.}
    \label{fig:critic-value}
\end{figure}

Figure \ref{fig:critic-value} shows the average long-term reward estimation of the CARL-EL algorithm under recommendations by real-time computation and by the cache, i.e., $Q_0$ and $Q_1$, respectively. It is shown that $Q_0$ is consistently larger than $Q_1$, showing that the model successfully learns the advantage of real-time recommendations over cached recommendations. Moreover, $Q_0$ and $Q_1$ have similar trendings over one day. For example, they both reach minimal values at about 12pm, while reaching maximal values at about 18pm. It can be explained by Eq. \eqref{eq:eigen-iterate}, where $\Lambda_a = Q_0 - Q_1 = V_0-V_1$, which means the difference between real-time and cached recommendations is independent of the cache ratios $D_0(t)$ and $D_1(t)$.

\begin{figure}
    \centering
    \includegraphics[width=\columnwidth]{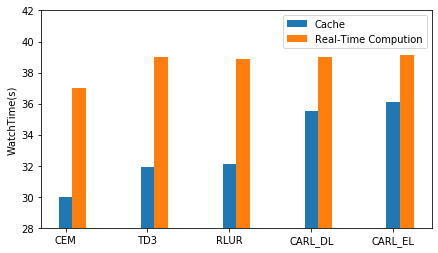}
    \caption{Average watch time per video of recommendations by real-time computation and by the cache.}
    \label{fig:cache_realtime}
\end{figure}

We further discuss the user feedback under recommendations by real-time computation and by the cache. Figure \ref{fig:cache_realtime} shows the average watch time per video in recommendations by real-time recommendations and by the cache. Although the performance of cached recommendations is lower than that of real-time recommendations, the CARL model increases the performance of cached recommendations more than that of real-time recommendations. Such results show that effective cache modeling helps reduce the performance gap between real-time and cached recommendations.

\subsubsection{Computational Burden}
We test the total time cost of the recommender system under different methods. TD3 increases the time cost by 0.524\% compared with CEM, while the difference in the time costs among these RL methods (TD3, RLUR, CARL-DL, and CARL-EL) is less than 0.1\% because we keep actors unchanged in the experiments. Moreover, to verify the trade-off of computational burden and the recommendation performance introduced by the result cache, we test the RLUR in an experimental group without a result cache, i.e. all the requests are performed by real-time computation. Such settings increase the daily watch time by 0.642\%, but with a significant increase of total time cost by 67.6\%. In contrast, CARL-EL increases the daily watch time by 0.391\% compared to RLUR with nearly no increase in the time cost, which shows the effectiveness of CARL-EL.

\section{Conclusion} \label{sec:conclusion}
This paper proposes the CARL model to reinforce the long-term rewards in the recommender systems with a result cache.  CARL uses a cache state to represent whether the recommender
system performs recommendations by real-time computation or by
the cache. The cache state is determined by the computational load
of the recommender system. Moreover, we propose an EL algorithm to learn independent critics for CARL to solve the critic dependency.
Experiments show that CARL can significantly improve the users’
engagement when considering the result cache. CARL has been
fully launched in Kwai app, serving over 100 million users.
\bibliographystyle{ACM-Reference-Format}
\balance
\bibliography{cache_rl}


\begin{thebibliography}{16}


\ifx \showCODEN    \undefined \def \showCODEN     #1{\unskip}     \fi
\ifx \showDOI      \undefined \def \showDOI       #1{#1}\fi
\ifx \showISBNx    \undefined \def \showISBNx     #1{\unskip}     \fi
\ifx \showISBNxiii \undefined \def \showISBNxiii  #1{\unskip}     \fi
\ifx \showISSN     \undefined \def \showISSN      #1{\unskip}     \fi
\ifx \showLCCN     \undefined \def \showLCCN      #1{\unskip}     \fi
\ifx \shownote     \undefined \def \shownote      #1{#1}          \fi
\ifx \showarticletitle \undefined \def \showarticletitle #1{#1}   \fi
\ifx \showURL      \undefined \def \showURL       {\relax}        \fi
\providecommand\bibfield[2]{#2}
\providecommand\bibinfo[2]{#2}
\providecommand\natexlab[1]{#1}
\providecommand\showeprint[2][]{arXiv:#2}

\bibitem[Cai et~al\mbox{.}(2023)]%
        {cai2023reinforcing}
\bibfield{author}{\bibinfo{person}{Qingpeng Cai}, \bibinfo{person}{Shuchang Liu}, \bibinfo{person}{Xueliang Wang}, \bibinfo{person}{Tianyou Zuo}, \bibinfo{person}{Wentao Xie}, \bibinfo{person}{Bin Yang}, \bibinfo{person}{Dong Zheng}, \bibinfo{person}{Peng Jiang}, {and} \bibinfo{person}{Kun Gai}.} \bibinfo{year}{2023}\natexlab{}.
\newblock \showarticletitle{Reinforcing User Retention in a Billion Scale Short Video Recommender System}.
\newblock \bibinfo{journal}{\emph{arXiv preprint arXiv:2302.01724}} (\bibinfo{year}{2023}).
\newblock


\bibitem[Chen et~al\mbox{.}(2019)]%
        {chen2019top}
\bibfield{author}{\bibinfo{person}{Minmin Chen}, \bibinfo{person}{Alex Beutel}, \bibinfo{person}{Paul Covington}, \bibinfo{person}{Sagar Jain}, \bibinfo{person}{Francois Belletti}, {and} \bibinfo{person}{Ed~H Chi}.} \bibinfo{year}{2019}\natexlab{}.
\newblock \showarticletitle{Top-k off-policy correction for a REINFORCE recommender system}. In \bibinfo{booktitle}{\emph{Proceedings of the Twelfth ACM International Conference on Web Search and Data Mining}}. \bibinfo{pages}{456--464}.
\newblock


\bibitem[Fujimoto et~al\mbox{.}(2018)]%
        {fujimoto2018addressing}
\bibfield{author}{\bibinfo{person}{Scott Fujimoto}, \bibinfo{person}{Herke Hoof}, {and} \bibinfo{person}{David Meger}.} \bibinfo{year}{2018}\natexlab{}.
\newblock \showarticletitle{Addressing function approximation error in actor-critic methods}. In \bibinfo{booktitle}{\emph{International conference on machine learning}}. PMLR, \bibinfo{pages}{1587--1596}.
\newblock


\bibitem[Gao et~al\mbox{.}(2022)]%
        {gao2022kuairec}
\bibfield{author}{\bibinfo{person}{Chongming Gao}, \bibinfo{person}{Shijun Li}, \bibinfo{person}{Wenqiang Lei}, \bibinfo{person}{Jiawei Chen}, \bibinfo{person}{Biao Li}, \bibinfo{person}{Peng Jiang}, \bibinfo{person}{Xiangnan He}, \bibinfo{person}{Jiaxin Mao}, {and} \bibinfo{person}{Tat-Seng Chua}.} \bibinfo{year}{2022}\natexlab{}.
\newblock \showarticletitle{KuaiRec: A fully-observed dataset and insights for evaluating recommender systems}. In \bibinfo{booktitle}{\emph{Proceedings of the 31st ACM International Conference on Information \& Knowledge Management}}. \bibinfo{pages}{540--550}.
\newblock


\bibitem[Giannakas et~al\mbox{.}(2018)]%
        {giannakas2018show}
\bibfield{author}{\bibinfo{person}{Theodoros Giannakas}, \bibinfo{person}{Pavlos Sermpezis}, {and} \bibinfo{person}{Thrasyvoulos Spyropoulos}.} \bibinfo{year}{2018}\natexlab{}.
\newblock \showarticletitle{Show me the cache: Optimizing cache-friendly recommendations for sequential content access}. In \bibinfo{booktitle}{\emph{2018 IEEE 19th International Symposium on" A World of Wireless, Mobile and Multimedia Networks"(WoWMoM)}}. IEEE, \bibinfo{pages}{14--22}.
\newblock


\bibitem[Jiang et~al\mbox{.}(2020)]%
        {jiang2020dcaf}
\bibfield{author}{\bibinfo{person}{Biye Jiang}, \bibinfo{person}{Pengye Zhang}, \bibinfo{person}{Rihan Chen}, \bibinfo{person}{Xinchen Luo}, \bibinfo{person}{Yin Yang}, \bibinfo{person}{Guan Wang}, \bibinfo{person}{Guorui Zhou}, \bibinfo{person}{Xiaoqiang Zhu}, {and} \bibinfo{person}{Kun Gai}.} \bibinfo{year}{2020}\natexlab{}.
\newblock \showarticletitle{DCAF: A Dynamic Computation Allocation Framework for Online Serving System}.
\newblock \bibinfo{journal}{\emph{arXiv preprint arXiv:2006.09684}} (\bibinfo{year}{2020}).
\newblock


\bibitem[Johnson et~al\mbox{.}(2019)]%
        {johnson2019billion}
\bibfield{author}{\bibinfo{person}{Jeff Johnson}, \bibinfo{person}{Matthijs Douze}, {and} \bibinfo{person}{Herv{\'e} J{\'e}gou}.} \bibinfo{year}{2019}\natexlab{}.
\newblock \showarticletitle{Billion-scale similarity search with gpus}.
\newblock \bibinfo{journal}{\emph{IEEE Transactions on Big Data}} \bibinfo{volume}{7}, \bibinfo{number}{3} (\bibinfo{year}{2019}), \bibinfo{pages}{535--547}.
\newblock


\bibitem[Lillicrap et~al\mbox{.}(2015)]%
        {lillicrap2015continuous}
\bibfield{author}{\bibinfo{person}{Timothy~P Lillicrap}, \bibinfo{person}{Jonathan~J Hunt}, \bibinfo{person}{Alexander Pritzel}, \bibinfo{person}{Nicolas Heess}, \bibinfo{person}{Tom Erez}, \bibinfo{person}{Yuval Tassa}, \bibinfo{person}{David Silver}, {and} \bibinfo{person}{Daan Wierstra}.} \bibinfo{year}{2015}\natexlab{}.
\newblock \showarticletitle{Continuous control with deep reinforcement learning}.
\newblock \bibinfo{journal}{\emph{arXiv preprint arXiv:1509.02971}} (\bibinfo{year}{2015}).
\newblock


\bibitem[Liu et~al\mbox{.}(2017)]%
        {liu2017cascade}
\bibfield{author}{\bibinfo{person}{Shichen Liu}, \bibinfo{person}{Fei Xiao}, \bibinfo{person}{Wenwu Ou}, {and} \bibinfo{person}{Luo Si}.} \bibinfo{year}{2017}\natexlab{}.
\newblock \showarticletitle{Cascade ranking for operational e-commerce search}. In \bibinfo{booktitle}{\emph{Proceedings of the 23rd ACM SIGKDD International Conference on Knowledge Discovery and Data Mining}}. \bibinfo{pages}{1557--1565}.
\newblock


\bibitem[Rubinstein and Kroese(2004)]%
        {rubinstein2004cross}
\bibfield{author}{\bibinfo{person}{Reuven~Y Rubinstein} {and} \bibinfo{person}{Dirk~P Kroese}.} \bibinfo{year}{2004}\natexlab{}.
\newblock \bibinfo{booktitle}{\emph{The cross-entropy method: a unified approach to combinatorial optimization, Monte-Carlo simulation, and machine learning}}. Vol.~\bibinfo{volume}{133}.
\newblock \bibinfo{publisher}{Springer}.
\newblock


\bibitem[Wang et~al\mbox{.}(2017)]%
        {wang2017caching}
\bibfield{author}{\bibinfo{person}{Yanfeng Wang}, \bibinfo{person}{Mingyang Ding}, \bibinfo{person}{Zhiyong Chen}, {and} \bibinfo{person}{Ling Luo}.} \bibinfo{year}{2017}\natexlab{}.
\newblock \showarticletitle{Caching placement with recommendation systems for cache-enabled mobile social networks}.
\newblock \bibinfo{journal}{\emph{IEEE Communications Letters}} \bibinfo{volume}{21}, \bibinfo{number}{10} (\bibinfo{year}{2017}), \bibinfo{pages}{2266--2269}.
\newblock


\bibitem[Wang et~al\mbox{.}(2020)]%
        {wang2020cold}
\bibfield{author}{\bibinfo{person}{Zhe Wang}, \bibinfo{person}{Liqin Zhao}, \bibinfo{person}{Biye Jiang}, \bibinfo{person}{Guorui Zhou}, \bibinfo{person}{Xiaoqiang Zhu}, {and} \bibinfo{person}{Kun Gai}.} \bibinfo{year}{2020}\natexlab{}.
\newblock \showarticletitle{Cold: Towards the next generation of pre-ranking system}.
\newblock \bibinfo{journal}{\emph{arXiv preprint arXiv:2007.16122}} (\bibinfo{year}{2020}).
\newblock


\bibitem[Xue et~al\mbox{.}(2022)]%
        {xue2022resact}
\bibfield{author}{\bibinfo{person}{Wanqi Xue}, \bibinfo{person}{Qingpeng Cai}, \bibinfo{person}{Ruohan Zhan}, \bibinfo{person}{Dong Zheng}, \bibinfo{person}{Peng Jiang}, {and} \bibinfo{person}{Bo An}.} \bibinfo{year}{2022}\natexlab{}.
\newblock \showarticletitle{ResAct: Reinforcing Long-term Engagement in Sequential Recommendation with Residual Actor}.
\newblock \bibinfo{journal}{\emph{arXiv preprint arXiv:2206.02620}} (\bibinfo{year}{2022}).
\newblock


\bibitem[Zhan et~al\mbox{.}(2022)]%
        {zhan2022deconfounding}
\bibfield{author}{\bibinfo{person}{Ruohan Zhan}, \bibinfo{person}{Changhua Pei}, \bibinfo{person}{Qiang Su}, \bibinfo{person}{Jianfeng Wen}, \bibinfo{person}{Xueliang Wang}, \bibinfo{person}{Guanyu Mu}, \bibinfo{person}{Dong Zheng}, \bibinfo{person}{Peng Jiang}, {and} \bibinfo{person}{Kun Gai}.} \bibinfo{year}{2022}\natexlab{}.
\newblock \showarticletitle{Deconfounding Duration Bias in Watch-time Prediction for Video Recommendation}. In \bibinfo{booktitle}{\emph{Proceedings of the 28th ACM SIGKDD Conference on Knowledge Discovery and Data Mining}}. \bibinfo{pages}{4472--4481}.
\newblock


\bibitem[Zhang et~al\mbox{.}(2024)]%
        {zhang2024unex}
\bibfield{author}{\bibinfo{person}{Gengrui Zhang}, \bibinfo{person}{Yao Wang}, \bibinfo{person}{Xiaoshuang Chen}, \bibinfo{person}{Hongyi Qian}, \bibinfo{person}{Kaiqiao Zhan}, {and} \bibinfo{person}{Ben Wang}.} \bibinfo{year}{2024}\natexlab{}.
\newblock \showarticletitle{UNEX-RL: Reinforcing Long-Term Rewards in Multi-Stage Recommender Systems with UNidirectional EXecution}.
\newblock \bibinfo{journal}{\emph{arXiv preprint arXiv:2401.06470}} (\bibinfo{year}{2024}).
\newblock


\bibitem[Zou et~al\mbox{.}(2019)]%
        {zou2019reinforcement}
\bibfield{author}{\bibinfo{person}{Lixin Zou}, \bibinfo{person}{Long Xia}, \bibinfo{person}{Zhuoye Ding}, \bibinfo{person}{Jiaxing Song}, \bibinfo{person}{Weidong Liu}, {and} \bibinfo{person}{Dawei Yin}.} \bibinfo{year}{2019}\natexlab{}.
\newblock \showarticletitle{Reinforcement learning to optimize long-term user engagement in recommender systems}. In \bibinfo{booktitle}{\emph{Proceedings of the 25th ACM SIGKDD International Conference on Knowledge Discovery \& Data Mining}}. \bibinfo{pages}{2810--2818}.
\newblock


\end{thebibliography}

\appendix

\section{Proof of Proposition \ref{prop:eigen}} \label{appendix:proof-eigen}

\begin{proof}
We consider the iterative function Eq. \eqref{eq:iteration} of the expected long-term rewards $L_0$ and $L_1$. By taking the difference between the first equation and the second equation in Eq. \eqref{eq:iteration}, we have $L_0 - L_1 = V_0 - V_1$, i.e. $\Lambda_0 = \Gamma_0$. Then, by taking a weighted summation of the first and second equations in Eq. \eqref{eq:iteration} with weight $D_0(t)$ and $D_1(t)$ respectively, we have
\begin{equation}
    \begin{aligned}
        &D_0(t)L_0(\boldsymbol{s}_t,\boldsymbol{a}_t) + D_1(t)L_1(\boldsymbol{s}_t) \\
        &=D_0(t)V_0(\boldsymbol{s}_t,\boldsymbol{a}_t) + D_1(t)V_1(\boldsymbol{s}_t) \\
        &+\gamma \left[D_0(t)+D_1(t)\right]D_0(t+1)\mathbb{E}_{\boldsymbol{a}_{t+1}}L_0\left(\boldsymbol{s}_{t+1},\boldsymbol{a}_{t+1}\right) \\
        &+\gamma \left[D_0(t)+D_1(t)\right]D_1(t+1)\mathbb{E}_{\boldsymbol{a}_{t+1}}L_1\left(\boldsymbol{s}_{t+1},\boldsymbol{a}_{t+1}\right)
    \end{aligned}
\end{equation}
given the fact that $D_0(t) + D_1(t) = 1$, we have
\begin{equation}
    \begin{aligned}
        &D_0(t)L_0(\boldsymbol{s}_t,\boldsymbol{a}_t) + D_1(t)L_1(\boldsymbol{s}_t) \\
        &=D_0(t)V_0(\boldsymbol{s}_t,\boldsymbol{a}_t) + D_1(t)V_1(\boldsymbol{s}_t) \\
        &+\gamma \mathbb{E}_{\boldsymbol{a}_{t+1}}\left[D_0(t+1)L_0\left(\boldsymbol{s}_{t+1},\boldsymbol{a}_{t+1}\right)+D_1(t+1)L_1\left(\boldsymbol{s}_{t+1},\boldsymbol{a}_{t+1}\right)\right]
    \end{aligned}
\end{equation}
which completes the proof.
\end{proof}

\end{document}